\documentclass[10pt, conference, compsocconf]{IEEEtran}
\usepackage{framed,multirow}
\usepackage{amssymb}
\usepackage{latexsym}
\usepackage{authblk}
\usepackage{url}
\usepackage{xcolor}
\definecolor{newcolor}{rgb}{.8,.349,.1}
\usepackage{caption}
\usepackage{subfig}
\usepackage{times}
\usepackage{epsfig}
\usepackage{graphicx}
\usepackage{amsmath}
\usepackage{amssymb}
\usepackage[pagebackref=true,breaklinks=true,letterpaper=true,colorlinks,bookmarks=false]{hyperref}
\begin{document}
\title{Beyond Visual Semantics: Exploring the Role of Scene Text in Image Understanding}

\author[1]{Arka Ujjal Dey}
\author[2]{Suman K. Ghosh}
\author[3]{Ernest Valveny}
\author[1]{Gaurav Harit}
\affil[1]{IIT Jodhpur, Rajasthan, India}
\affil[3]{Computer Vision Center, Universitat Aut\`{o}noma de Barcelona, Bellaterra (Barcelona)}
\affil[2]{University of Rouen, France}

\maketitle

\begin{abstract}

Images with visual and scene text content are ubiquitous in everyday life. However, current image interpretation systems are mostly limited to using only the visual features, neglecting to leverage the scene text content. In this paper, we propose to jointly use scene text and visual channels for robust semantic interpretation of images. We do not only extract and encode visual and scene text cues, but also model their interplay to generate a contextual joint embedding with richer semantics. The contextual embedding thus generated is applied to retrieval and classification tasks on multimedia images, with scene text content,  to demonstrate its effectiveness. In the retrieval framework, we augment our learned text-visual semantic representation with scene text cues, to mitigate vocabulary misses that may have occurred during the semantic embedding. To deal with irrelevant or erroneous recognition of scene text, we also apply query-based attention to our text channel. We show how the multi-channel approach, involving visual semantics and scene text, improves upon state of the art.
 
\end{abstract}

\section{Introduction}\label{sec:intro}
Images are the prevalent choice of expression these days, as they are often more engaging and less intrusive than other media. Often images use embedded scene text, in addition to visual elements, to express ideas more lucidly. Such images with visual and embedded scene text are ubiquitous in everyday life, in the form of printed advertisements, posters, propaganda bills, storefront views, and similar variants. The scene text content in such images is often crucial in the interpretation of the image. More importantly the scene text along with the visual contents often provide useful context to understand these media.

Text detection and recognition frameworks have matured in recent times, providing appealing results \cite{t_jaderberg2014synthetic,t_liu2018fots} while handling real life scenarios like complex backgrounds \cite{t_liao2017textboxes,t_jaderberg2016reading}, irregular font sizes or arbitrarily oriented text \cite{t_liu2018fots}. 
With these advances, the underlying scene text in images, which has been inaccessible until now in most image understanding tasks, can now be leveraged to interpret images in a more generalized way.
However, the use of scene text in image understanding thus far, has been scarce and constrained, basically to the realm of fine-grained classification tasks \cite{me_bai2017integrating,me_karaoglu2017text,me_karaoglu2017words,me_karaoglu2013text} and more recently to Visual Question Answering (VQA) \cite{tmg_singh2019textvqa,tmg_biten2019scene}. However, these works treat visual and text features as separate channels and do not model the semantic relationships between them. 

\begin{figure}
\vspace{-15pt}
\centering
\subfloat[]{\label{imgeg}\includegraphics[scale=0.30]{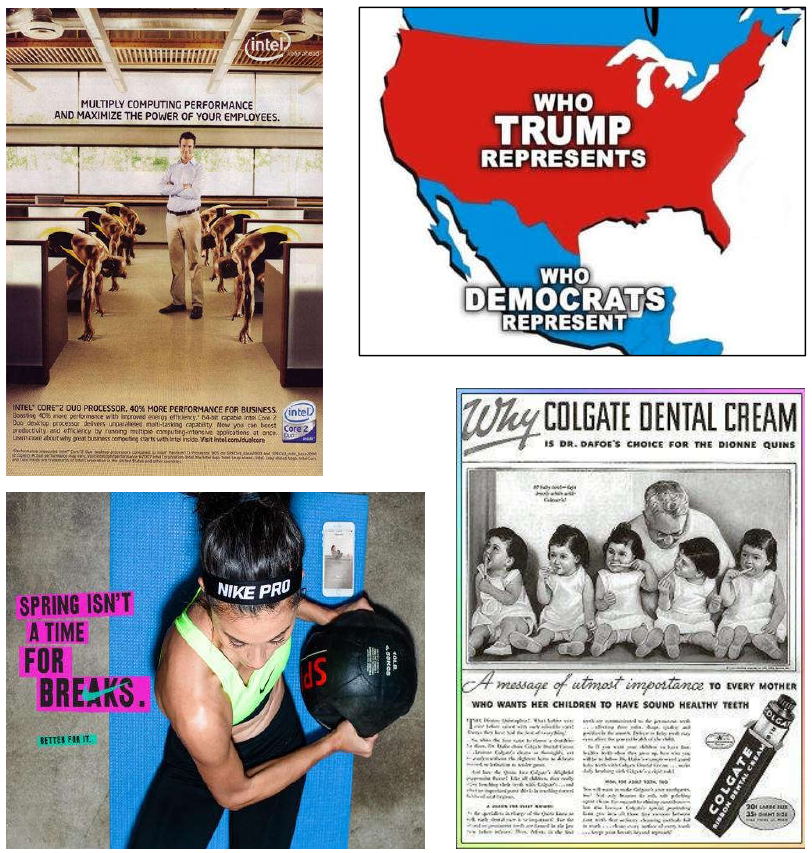}}
\subfloat[]{\label{basic_sot}\includegraphics[scale=0.51]{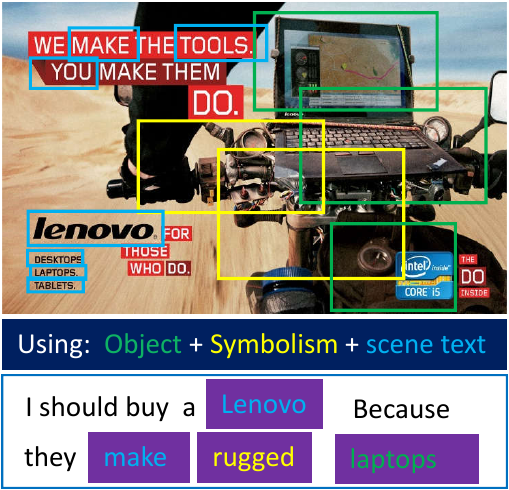}}
\vspace{-10pt}
\caption{(a) Complementary nature of text and visual cues: In some cases the visuals can be symbolic, but embedded text gives away the context[top-left, top-right], in other cases the visuals can be simple to understand but the text can be obtuse[bottom-left]. Further, the amount of text content can vary widely [top-right, bottom-right] 
(b) Basic idea: Use detected visual symbolism and objects, together with scene text to reason about images}
\end{figure}

In this work, we go beyond the detection of text and visual objects by learning a joint contextual semantic embedding that aims at capturing the inter-object dynamics. This interaction is modelled using a Text-Visual graph and a Graph Attention network \cite{gat_velivckovic2017graph} to generate the final embedding. The inter-object relationships, along with the encoded features, augments the ability to reason about images. In order to show how the contextual semantic embedding can be adapted to different scenarios we apply the model in two datasets where context plays a critical role: advertisement images\cite{ad_hussain2017automatic} and tweets\cite{gomez2019exploring}. We address two different tasks on the Ad dataset\cite{ad_hussain2017automatic} (retrieval of relevant statements and topic classification), as well as a binary sentiment classification task (hate speech detection) on the tweet dataset\cite{gomez2019exploring}. Both datasets contain images where text, as well as visual elements, are purposefully used to propagate an agenda, a marketing strategy, or a hateful message as illustrated in Fig.\ref{imgeg}. They may also contain socio-cultural references, symbolism\cite{ad_ye2017advise,ad_hussain2017automatic} along with wit and humor. Reasoning about such images involves understanding the context and the relationship between all the elements in that context\cite{equalnotsame}. 

In summary, the main contributions of this work are the following: first, we model the interplay between the detected text and the visual cues to generate a contextual embedding that encodes the inter-object dynamics using an attentional relationship graph. Second, we show through experiments that this model can be applied successfully, improving state of the art, to different image understanding tasks, such as semantic retrieval of relevant statements, topic classification or sentiment classification. Third, we make additional contributions to better leverage the use of scene text for the specific task of retrieving relevant statements in the ad dataset: we propose a novel use of scene text by using both semantic and lexical information  
and we leverage the language structure of the statement by partitioning it into an action-reason pair to better model the relation between the semantics of the query and the image.
\section{Related works}\label{sec:relatedworks}

\subsection{ Use of scene text }\label{sec:img_und_sct} 

Using scene text for image understanding has been attempted mainly in the context of fine-grained image classification tasks \cite{me_bai2017integrating,me_karaoglu2017text,me_karaoglu2017words,me_karaoglu2013text}. Leveraging scene text present in an image may lead to better classification accuracy for certain types of images, e.g., storefront images \cite{me_karaoglu2017text,me_karaoglu2017words}. While in \cite{me_karaoglu2013text}, the authors use a spatial encoding of n-grams as text features, in \cite{me_karaoglu2017words} they argue for word-level features and use a vocabulary based Bag of Words (BOW) representation. In both works, the final representation is a combination of visual and text cues, without using any semantic information or modelling of the interaction between the text and visual cues. Only in \cite{me_bai2017integrating}, the authors proposed encoding the text using a semantic embedding\cite{tw_mikolov2013efficient} improving upon the previous results. 

Recently, motivated for last advances in scene text extraction, there is a surge of interest in systems and datasets that leverage scene text along with traditional visual cues, for instance in advertisement understanding\cite{ad_hussain2017automatic}, hate speech detection\cite{gomez2019exploring} or VQA\cite{tmg_singh2019textvqa,tmg_biten2019scene}. In these cases it is observed 
that visual features alone are not enough and extracting the scene text and encoding the context is critical for successful interpretation. Scene text features have been shown to be quite discriminative by themselves for advertisement understanding, as noted by the \href{https://people.cs.pitt.edu/~kovashka/ads\_workshop/}{CVPR AD Workshop} Challenge winners. We will rely on these results and we will also use a separate text channel encoding scene text semantics for ad retrieval. In the case of VQA the proposed baselines, built upon traditional VQA systems, combine the text channels along with the standard visual channel, but without trying to model the relationship between them. In our work, we will show that modeling such relationships generates rich contextual features leading to improved semantics.



\subsection{ Text and vision }
\label{sec:textandvision}
Language and vision are the two most important ways we communicate. Thus, their combination poses important challenges like image captioning \cite{BAI2018291}, text-based image retrieval (e.g., google image search) and Visual Question Answering \cite{mf_fukui2016multimodal} among others.
In most cases, the semantic encoding of the text, is used either in conjunction with visual features through fusion\cite{mf_fukui2016multimodal,mf_ben2017mutan} for VQA tasks, or it is used to define a common subspace\cite{me_faghri2017vse++} to project the visual representation into for retrieval tasks.  In \cite{ad_ye2017advise}, an embedding scheme projects images and statements into a common subspace, where retrieval is feasible. The embedding scheme used features from salient regions proposed by symbol detectors and automated captions generated by Densecap\cite{ac_johnson2016densecap}. The generated caption acted as external knowledge and was encoded with word-embedding\cite{tw_mikolov2013efficient}. 
The success of such methods\cite{ad_ye2017advise,twe_frome2013devise} in embedding visual features into a common semantic subspace can be largely attributed to the discriminative nature of text semantics\cite{tw_mikolov2013efficient,tw_pennington2014glove} facilitated by availability of huge text corpus.

In these works, the text originates from an external source (question, caption, annotation) and not in the form of scene text present within the image. While images may usually contain visual objects, symbolism, and motifs the advertisement and tweet images that we analyse in this work often use scene text content to drive home a clear message. Thus, while we find several related work exploring the visual symbolism present\cite{ad_ye2017advise,ad_doshisymbolic}, or attending\cite{ad_ahuja2018understanding} to different visual components, we also explore the role of scene text in conveying that take away message. 
\subsection{ Contextual Encoding }\label{sec:txtvisrel}

Given the nature of high-level tasks like VQA or captioning, both textual and visual cues convey essential contextual information to be leveraged. While feature fusion\cite{mf_fukui2016multimodal,mf_ben2017mutan} is a standard scheme for image representation encoding different modalities, it is preceded by feature aggregation of respective modalities. However, simple aggregation of local features from different modalities leads to loss of fine-grained spatial and contextual information that can be beneficial for high-level downstream tasks. Here is where attention\cite{Mnih:2014:RMV:2969033.2969073}, or relevance of the different detected components, comes into play. 
In the case of advertisement images, for instance, the need to attend to relevant information has found expression in various works\cite{ad_ahuja2018understanding,ad_ye2017advise,ad_doshisymbolic}. In these works we see examples of top-down attention, guided by the final task\cite{ad_ye2017advise} or linguistic cues, in the form of the text statement \cite{ad_doshisymbolic}, to attend to the visual features corresponding to symbolism and objects. Recently, attention on graphs describing the relations among different components has been proposed with the Graph Attentional Layer (GAT)\cite{gat_velivckovic2017graph,twe_vashishth2019incorporating,vr_li2019relation} to encode the context. We will leverage GAT in our contextual encoder as a way to explore the interplay between the detected textual and visual local features, and encode their contextual information, to generate rich semantics. 

\section{Method}
\begin{figure*}
\begin{center}
\includegraphics[width=0.9\textwidth]{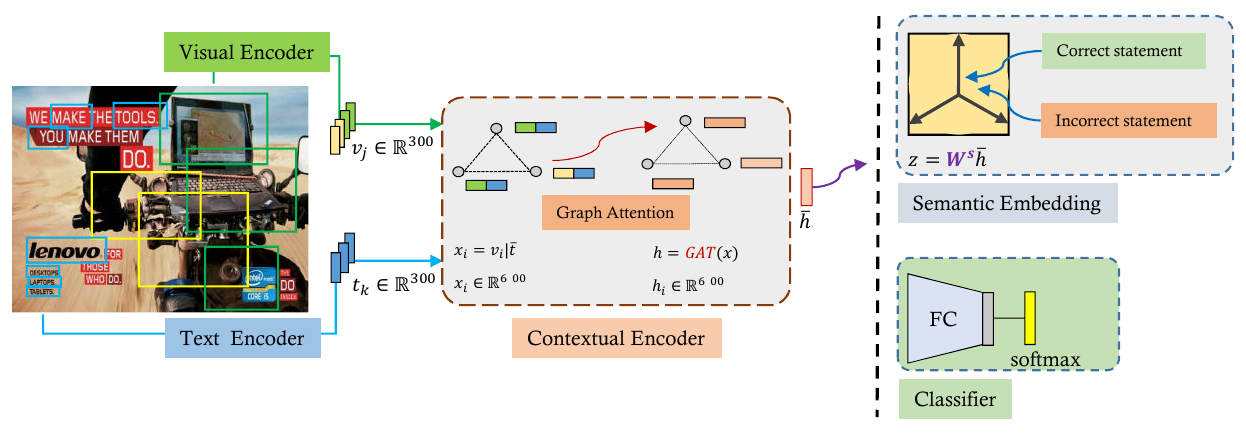}
\caption{ Model architecture of the Proposed Contextual Embedding applied to the separate tasks of semantic embedding and classification }\label{fig:vsem_main}
\end{center}
\end{figure*}
Fig.\ref{fig:vsem_main} gives a detailed illustration of our proposed model, which extracts and encodes visual and scene text cues to generate a contextual encoding, applicable to different tasks, viz semantic embedding, classification. The basic stages of our pipeline consist of a Visual Encoder, a Scene Text Encoder, and a Multi-Modal Contextual encoder.

\subsection{Visual Encoder}
Images often consists of multiple visual elements providing different semantic information. Thus we argue that local region patches are better suited\cite{me_karaoglu2017words} to this task than global image-level features.
We use two different channels to generate meaningful local patches. We use a pre-trained standard object detector\cite{c_huang2017speed}, to detect salient objects in the image, which can convey relevant semantic information. Recently it has been shown that symbolism associated with the local visual patches in the image (for instance, concepts like danger, cool, freedom) can play a significant role in semantic understanding \cite{ad_ye2017advise}. Thus, we also leverage symbol annotations in the dataset\cite{ad_hussain2017automatic} and use a pre-trained \cite{ad_ye2017advise} symbolism detector to generate an additional set of local region patches. A pre-trained deep network \cite{c_he2016deep} is used to extract visual features $v_i$ $\in$ $R^{300}$, corresponding to both detected object and symbolism local patches. 

\subsection{Scene Text Encoder}\label{sec:scenetxtenc}
While scene text (present in the image) may be a rich source of information for semantic understanding of the image, extracting that text involves dealing with complexities like cluttered background, orientation, or uneven lighting. 
Considering that the accuracy of text extraction is a critical factor for later image reasoning, we have analyzed different OCR alternatives\cite{t_liao2017textboxes,t_jaderberg2016reading} to evaluate their impact on the final system. Finally, we settled on using Google Vision API, as it leads to improved text extraction, generating legible scene text for about $94\%$ images. 
The extracted text is embedded~\cite{tw_mikolov2013efficient} into a word embedding space that encodes the semantics of the text.

\paragraph{Anchor based Text Attention}
The number of recognized scene text words varies widely and besides, not all of the words are relevant to the given task. Therefore, we propose to encode the detected words in terms of a fixed number of anchors (or clusters) specific for every task. In the case of the tasks in the Ad dataset, the anchors are the 15 statements associated with each image (see section \ref{sec:task1} for details). For the Tweet dataset, we use the same from \url{hatebase.org}, as used for dataset collection\cite{gomez2019exploring}. 

Thus given $n$ recognized scene text words $[t^{'}_1.. t^{'}_n]$, we encode them as $[t_{1} ..t_{k}]$, in terms of the $k$ task dependent anchors $A_k$:  
\begin{equation}\label{eqn:textattention}
t_{k}= \sum_{i=0}^{n} r_{i,k} t^{'}_i \text{, where } r_{i,k}= \sum_{j}\frac{1}{1 + d(t^{'}_i,A_{k,j})} 
\vspace{-5pt}
\end{equation}
\noindent{$r_{i,k}$ gives the similarity between a scene text word $t^{'}_i$ and anchor $A_k$, based on distance measure $d$. When an anchor has multiple words (statements in the ad dataset), $r_{i,k}$ is the sum of similarities with all anchor words $A_{k,j}$. Thus, given a variable number of detected words, only those that are similar to the anchors are considered relevant.}

\subsection{Contextual VT (Visual Text) Encoder}\label{sec:tvre}
One of our main contributions is a representation that captures the rich interplay between the text and visual cues. Such a representation entails a) defining a compositional strategy encoding the contextual relationships among the co-occurring text and visual features, and b) capturing their interaction. 

\paragraph{Compositional Strategy: Text Visual Graph}
In the case of text semantics, the strategy employed to encode context is usually sequential \cite{tw_mikolov2013efficient} characterized by a sliding window. Our recognized text and visual objects do not have any particular sequence ordering, and thus we take inspiration from the recently proposed graph-structured context\cite{twe_vashishth2019incorporating}. 

The top 10 visual objects ${v_j}$ detected by the visual encoder and the $k$ task dependent text anchors ${t_k}$ provided by the scene text encoder are represented as nodes to construct a fully connected graph $G=(V,E)$, with $V=v \cup t$. 
However, the text and visual nodes have features from different domains and are not directly comparable. Thus, we augment visual nodes with the mean of their adjacent text nodes, and similarly for text nodes we augment them with the mean of their adjacent visual nodes:

\begin{equation}
 x_i=v_i || \Bar{t}
 \quad\mathrm{or}\quad 
 x_i=\Bar{v}||t_i 
\end{equation}

We assume the graph is fully connected based on our earlier hypothesis of relatedness between all objects in the image (text and visual). The edge weights representing the relevance between two nodes are implicitly learned through the Graph Attention Layer.

\paragraph{Interaction scheme: Relation Encoder}
We model the text-visual interplay in our relationship encoder by allowing attentional interaction amongst the nodes of the graph through a Graph Attention Layer \cite{gat_velivckovic2017graph}. We allow nodes in a similar context, in this case an image, to influence each other's representation. Our interaction scheme is similar to the \textit{Implicit Relational Encoder} proposed in \cite{vr_li2019relation}, but we differ in our design of the attention mechanism and also allow for multimodal text-visual interaction. 
Given the input features $x_i$ of a node, we learn a shared projection matrix $W$, and perform self-attention on the adjacent nodes to generate the output feature $h_i$ for that node 

\begin{equation}
 h_i=\sum_{j} \alpha_{ij}.W x_i
\quad\mathrm{with }\quad 
 \alpha_{ij}= softmax(e_{i,j}) 
\end{equation}

\noindent{where $\alpha_{ij}$ is the attention weight defined using $e_{i,j}$ representing the importance of node $j$ to node $i$. It is computed by a single layer feed forward network akin to \cite{gat_velivckovic2017graph}. We define the final aggregated contextual feature as $\Bar{h}$, the mean of all the nodes.}

\section{Application of the Model}

\begin{figure}
\includegraphics[width=80mm, height=25mm ]{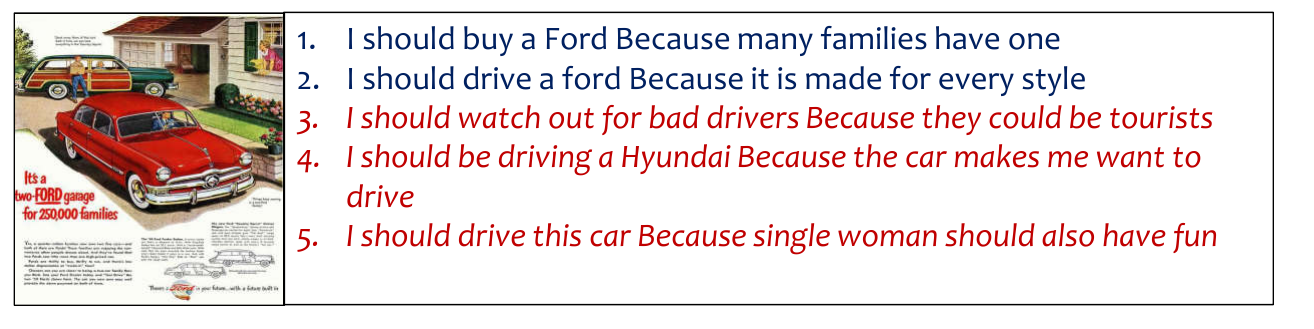}
\caption{ A sample Ad image, with relevant sentences in blue and irrelevant sentences red. The task is to rank the relevant sentences ahead of the irrelevant ones, given an example Ad image. Showing only 5 of 15 statements for brevity }\label{basicar}
\end{figure}

\subsection{Task 1: Image-Statement Relevance }
\label{sec:task1}
For this task we will use the Ad image dataset introduced in \cite{ad_hussain2017automatic}. In a later work \cite{ad_ye2017advise}, a retrieval task was proposed, where the goal is to match an Ad image against relevant sentences. For each image 3 relevant and 12 non-relevant sentences are provided. See Fig.\ref{basicar} for an example of this task.

\subsubsection*{Learning of the contextual semantic embedding}
The Image-Statement Relevance task entails matching statements against images. Thus, we need the image and statement representations to be comparable by a distance. As explained in section \ref{sec:tvre}, images are are represented by the aggregate of their contextual embedding given by $\Bar{h}$. Statements are encoded as the aggregate of their word2vec word embeddings. To make them comparable we project the aggregated contextual vector $\Bar{h}$ into a semantic space $z$, where matching against relevant statements is feasible, as depicted in Fig.\ref{fig:vsem_main}. The weights $W^{s}$ of the projection matrix are learned through triplet training with relevant and irrelevant statements, minimizing this triplet loss 
\begin{equation}
\label{eqn:loss}
l(z,s,\theta)= \sum_{i=1}^{B} \sum_{j \in ns(i)} \| z_i - s_i\| - \| z_i - s_j\| + \beta 
\end{equation}
where B is the batch size, $\beta$ is the margin of triplet loss and $z$ is the semantic embedding, with $s_i$ and $s_j$ being the randomly sampled positive and negative statement semantic features respectively. Given the $i^{th}$ image, $ns(i)$ denotes the set of irrelevant statement $s_j$ is sampled from. 

\subsubsection*{Retrieval framework} 
\begin{figure}
\includegraphics[width=80mm, height=60mm ]{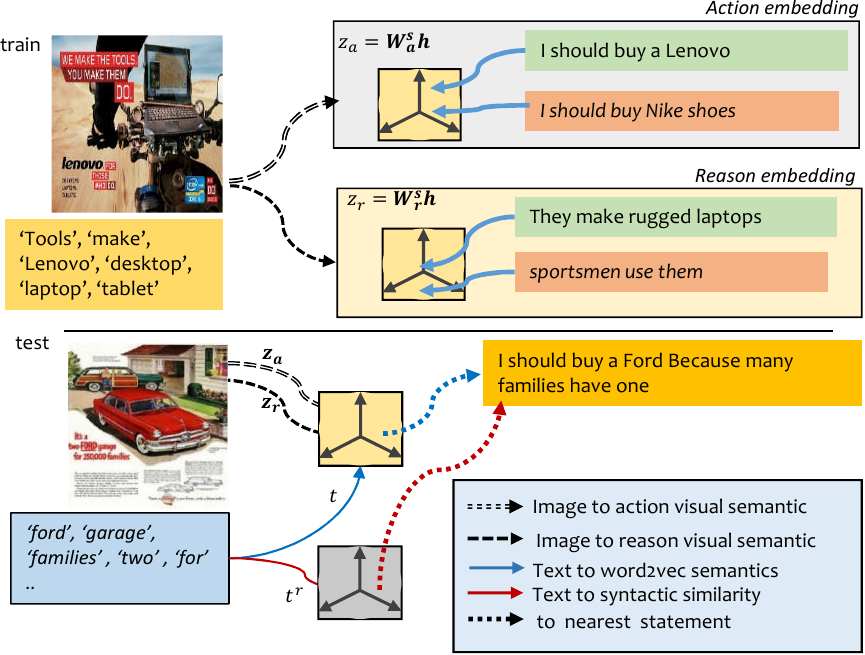}
\caption{ Sentence Relevance Task: Training, Testing }\label{fig:vsem_tt}
\end{figure}

We build upon the semantic contextual embedding learnes as explained in the previous section to define the complete framework for the Image-Statement Relevance task shown in Fig.\ref{fig:vsem_tt}. We integrate some specific model components that leverage certain properties of the advertisement images to boost performance in this task. More specifically, in training we learn separate semantics based on statement action-reason partitioning. During testing, we integrate additional text channels component to mitigate vocabulary misses. 

\paragraph{Partitioning}
Analogous to some recent works
\cite{DBLP:journals/corr/abs-1809-08697,DBLP:journals/corr/AndreasRDK16} that combine natural language and vision we take advantage of the linguistic structure of the statements. The statements in this task can be \emph{partitioned} into a couplet of action and reason : ``I should $<action>$ because $<reason>$." e.g., ``I should buy a Lenovo because they are rugged laptops", as can be seen in Fig. \ref{basic_sot}. We exploit this structure and learn, using the triplet loss defined in eq.\ref{eqn:loss}, two separate semantic embeddings, viz one related to actions $z_a$, and another one to reasons $z_r$. Thus, given an image, we now evaluate its relevance separately for the action and the reason. Such partitioning allows not just exploiting fine-grained intermediate data, but it also enables to mitigate long term dependencies associated with long sentences \cite{p_niu2017hierarchical}.

\paragraph{Scene Text Semantic Channel}
During test time, while matching images against statements, we augment the trained contextual semantic embedding with an independent scene text semantics channel. This text semantics uses the same anchor based encoding of the scene text as described in eq. \ref{eqn:textattention}, but in this case we only use the query statement as a single anchor. 

\paragraph{Lexical similarity scoring}
Often brand names, or brand-related terms, like `googling, Mcchicken', are not present in the pretrained word2vec vocabulary used for text semantics. This can be further aggravated by erroneous word recognition. In particular, for the Ad dataset, a total of $15\%$ of the 3 million recognized words could not be mapped to any semantic vector. One way to cope with vocabulary misses in word embedding is to use Lexical Similarity. It provides us a way to check for the similarity between the raw scene text and the query statement without the need for any further embedding. This enables us to use all the extracted scene text words, taking advantage of any word correspondence. We measure the lexical distance $d({s_j}^r,t^r)$, as the cosine distance between tf-idf vectors of the raw scene text words ${t_i}^r$ and the query statement words ${{s_j}_k}^r$.
\vspace{-5pt}
\paragraph{Final matching: Combining contextual semantics with text scoring}
The final distance measure used for the ranking of the query statement tries to capture the semantic distance between the statement $s_j$ and the image taking into account the contextual semantic features $z$ (with action-reason partitioning $z_a$ and $z_r$), the semantic text features $t$, and the lexical distance from the scene text features $t^r$. It is given by:

\vspace{-15pt}
\begin{equation}\label{eqn:finalloss}
\underset{j \in Q } \arg \min d(z_a,s_{j_a})+ d(z_r,s_{j_r})+ d(t,s_j)+ d(t^r,{s_j}^r) 
\end{equation}
where we have $Q$ query statements to rank against an image. 

\vspace{-5pt}
\subsection{Task 2: Classification}
We also apply the contextual semantic embedding to two different classification tasks: topic classification on the Ad Image dataset\cite{ad_hussain2017automatic}  and tweet classification on the MMHS150K dataset\cite{gomez2019exploring}. While the Topic classification\cite{ad_hussain2017automatic} task consists of categorizing the Ad image under one of 38 different product heads viz, {`car',`beauty',`coffee'}, the tweet classification  involves marking tweets as hateful or benign. 
In both cases the classifier is built upon the contextual encoding framework as depicted in Fig.\ref{fig:vsem_main}.
In particular, the contextual representation is fed to a softmax classifier and trained end-to-end with cross-entropy loss.

\section{Experimental Results}
\subsection{Task1: Image-Statement Relevance}
For the Image-Statement Relevance task we follow the protocol introduced in the \href{https://people.cs.pitt.edu/~kovashka/ads\_workshop/}{CVPR Workshop}
\footnote{https://people.cs.pitt.edu/~kovashka/ads\_workshop/}, 
and rank 15 statements (3 relevant, 12 non relevant) based on their relevance or similarity to the image.

\vspace{-5pt}
\paragraph{Metrics} We compute 3 different metrics: 1) Accuracy, which records a hit whenever any of the 3 relevant statements is picked 2) Rank Average, which is the average rank of the highest-ranked relevant statement and 3) Recall at 3, which denotes the number of correct statements ranked in top 3. For a good model, we expect high accuracy and recall, with a low average rank. 

\vspace{-5pt}
\subsubsection{ Comparison with the state-of-the-art}\label{sec:soa}
In this section, we compare our results with the current state of the art. We first give a brief description of the methods used for comparison. VSE++ \cite{me_faghri2017vse++} is one of the major visual semantic embedding schemes, but it does not incorporate the symbolism or scene text content present in the Ad image. ADVISE \cite{ad_ye2017advise} played the crucial role of leveraging the symbol annotation\cite{ad_hussain2017automatic} present in the dataset, and use the symbol channel in the visual semantic embedding. While these schemes do use external knowledge, in the form of automatically generated captions\cite{ac_johnson2016densecap}, to augment their visual understanding, we are the first ones to formally introduce scene text in the context of visual understanding. Both VSE++ \cite{me_faghri2017vse++} and ADVISE \cite{ad_ye2017advise} had also participated in the CVPR 2018 Workshop Challenge, organised on this dataset.  In the results we can clearly see the improvement brought upon by our complete framework using contextual semantics trained on visual and text features, augmented with text scoring and statement partitioning.
\begin{table}[h]
 
\begin{center}
\small
\caption{Comparison with state-of the-art. Results marked with * do not use the exactly our same partitions for training and test. }\label{result_sem_comp}
 \begin{tabular}{l|c|c }
 Model & RankAvg $\downarrow$ & Accuracy $\uparrow$ \\ \hline
VSE++ \cite{me_faghri2017vse++} & 3.85 & 66.6 *\\ 
ADVISE \cite{ad_ye2017advise} & 3.55 & 72.84 * \\ 
CVPRW winner & - & 82 * \\
\bf{Our full system} & 3.09 & \bf{90.9} \\
 \end{tabular}
\end{center} 
\end{table}

\vspace{-15pt}
\subsubsection{Ablation Study}
\paragraph{Training of Semantic Contextual Embedding}
In Tab.\ref{abl_sem} we analyze the contribution of the different channels and components involved in training the semantic contextual embedding. 
Visual and text baselines are proposed to show the contribution of each individual channel in the contextual embedding (columns 1 and 2 in Tab.\ref{abl_sem}).
The visual baseline only uses the ResNet visual features and excludes the use of scene text or GAT in the pipeline. For the  text baseline, only the scene text word embeddings are used. In both cases, the local features are aggregated to generate a semantic vector trained with triplet loss. In columns 3 and 4 textual and visual features are fused with simple concatenation while the full model using the contextual VT encoder is shown in column 5.

\vspace{-5pt}
\paragraph{Contribution of the different channels}
For the sentence relevance task, as eq.\ref{eqn:finalloss} shows, the ranking involves, not just the visual semantic features, but also semantic and lexical text features. In Tab.\ref{tab:ab_sentrev}, we detail the contribution of each of these separate channels. We also show the improvement due to using statement based attention when aggregating scene text features.

\begin{table}[h!]
 \begin{center}
 \caption{Semantic Embedding : Role of Text and Visual channels, partitioning and Contextual VT encoder in semantic embedding }
 \label{abl_sem}
 \begin{tabular}{l|r|r|r|r|r}
 Visual & $$\checkmark$$ & $\times$ & $\checkmark$ & $\checkmark$ & $\checkmark$ \\
 Text & $\times$ & $\checkmark$ & $\checkmark$ & $\checkmark$ & $\checkmark$ \\
 Partitioning & $\times$ & $\times$ & $\times$ & $\checkmark$ & $\checkmark$ \\ 
 Contextual VT Encoder & $\times$ & $\times$ & $\times$ &$\times$ &$\checkmark$ \\ \hline
 Accuracy $\uparrow$ & 55.6 &74.4 &82.4 &83.5 &85.7 \\ 
 RankAvg $\downarrow$ &4.77 & 4.31 &3.34 &3.29 &3.2 \\ 
 Recall@3 $\uparrow$ &1.4 & 1.7 & 2.12 &2.14 &2.19 \\
 \end{tabular}
 \end{center}
\end{table}

\begin{table}[h!]
 \begin{center}
 \caption{Sentence Relevance: Role of components Semantic and Text channel in Sentence Relevance task}
 \label{tab:ab_sentrev}
 \begin{tabular}{l|r|r|r|r}
  Text Semantic & $\times$ & $\checkmark$ & $\times$ & $\times$ \\
 Text Semantic w/ attention & $\times$ & $\times$ & $\checkmark$ & $\checkmark$ \\ \hline
 Lexical & $\times$ & $\times$ & $\times$ & $\checkmark$ \\
 Semantic Embedding & $\checkmark$ & $\times$ & $\times$ & $\checkmark$ \\ \hline
 Accuracy $\uparrow$ & 85.76 &72 &74.4 & 90.9 \\
 RankAvg $\downarrow$ & 3.2 &4.52 & 4.31 & 3.08 \\ 
 Recall@3 $\uparrow$ & 2.19 & 1.6 & 1.7 & 2.3 \\
  \end{tabular}
  \vspace{-10pt}
 \end{center}
\end{table}

\subsubsection{Qualitative Results}
\begin{figure}
\begin{center}
\includegraphics[width=80mm, ]{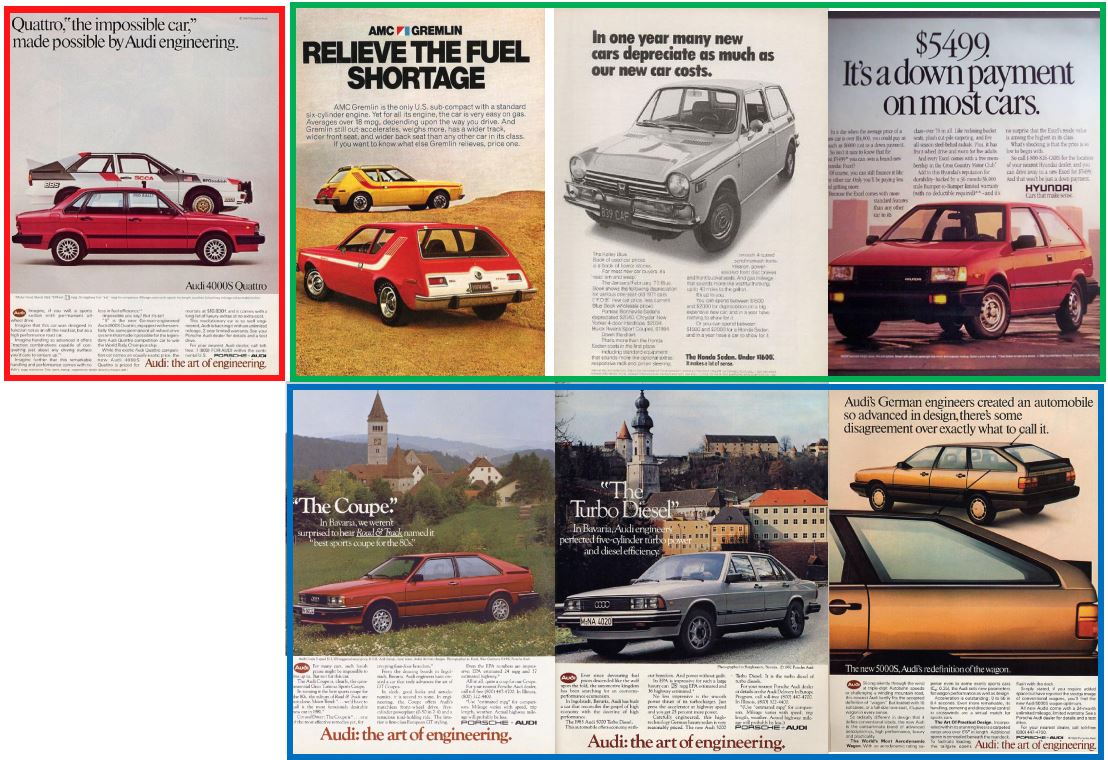}
\caption{ Semantic Retrieval. The semantic features of the query image, bounded in red, is used to find its top 3 matches among the other test images. The Top row lists images, that were retrieved using  Visual cues based semantic feature. The bottom row uses our contextual encoder for semantic features, using both scene text and visual cues. Our improvements leads to retrieval of images not just pertaining to cars, but also gets the type and brand right. }\label{abl_vistxtmain}
\end{center}
\end{figure}

Fig.\ref{abl_vistxtmain} shows examples of query by image, i.e. the semantic features of an image are used to find similar images. We show that the proposed scheme can encode the visual and text cues, and generate a holistic semantic feature. Comparison with the baseline that uses only visual features shows the effectiveness of scene text in generating more fine-grained results. \\

\begin{figure*}
\centering
\includegraphics[scale=0.45]{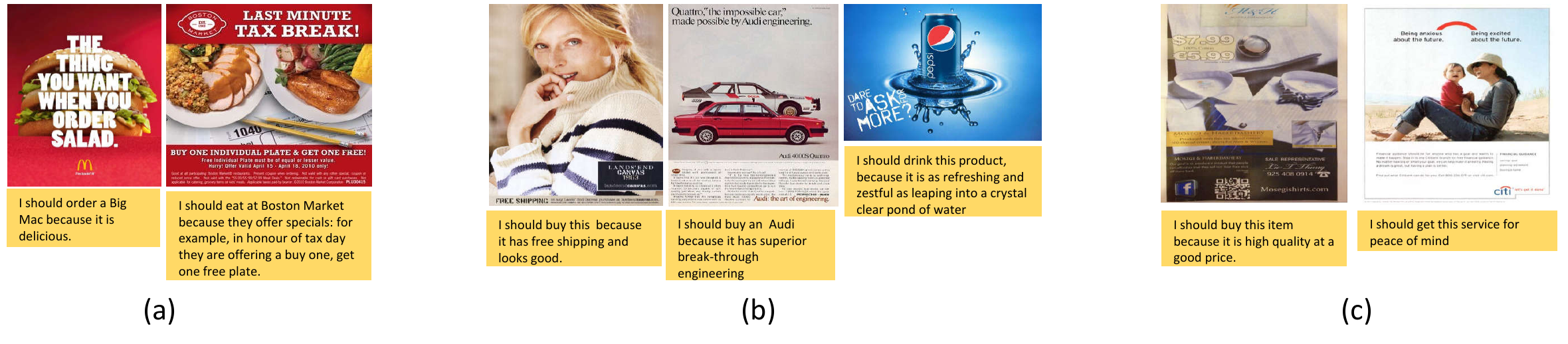}
\vspace{-5pt}
\caption[Ablated Instances]{(a) Examples where just visual features were not enough to generate robust semantics, and only with the incorporating of Scene Text content in Semantic Embedding were we able to map the images to their relevant statements.
(b)Examples where semantic features with visual and text, still fell short of retrieving the correct relevant statements and Partitioning had to help.
(c)Examples that required exploiting Non-Literal relationships using our visual text relationship encoder. }\label{fig:abl_all}
\end{figure*}

In Fig.\ref{fig:abl_all}a, we display instances where the visual features by themselves were not able to map the image to the correct statements, and we had to incorporate scene text in the semantic representation. This can be attributed to the co-occurrence of certain semantically related words in both the scene text and the relevant statements. However, the simple co-occurrence of semantically related words does not suffice for all examples, as is illustrated in Figs.\ref{fig:abl_all}b and \ref{fig:abl_all}c. In particular, in Fig.\ref{fig:abl_all}c, we show test instances that were only correctly mapped to their statements when we incorporated the relationship encoder, going beyond simple visual or text similarity and exploiting non-literal relationships. For example, in the second example, it had to relate that getting the service amounted to bridging the challenge between being anxious and excited. 

\begin{figure}
\begin{center}
\includegraphics[scale=0.70]{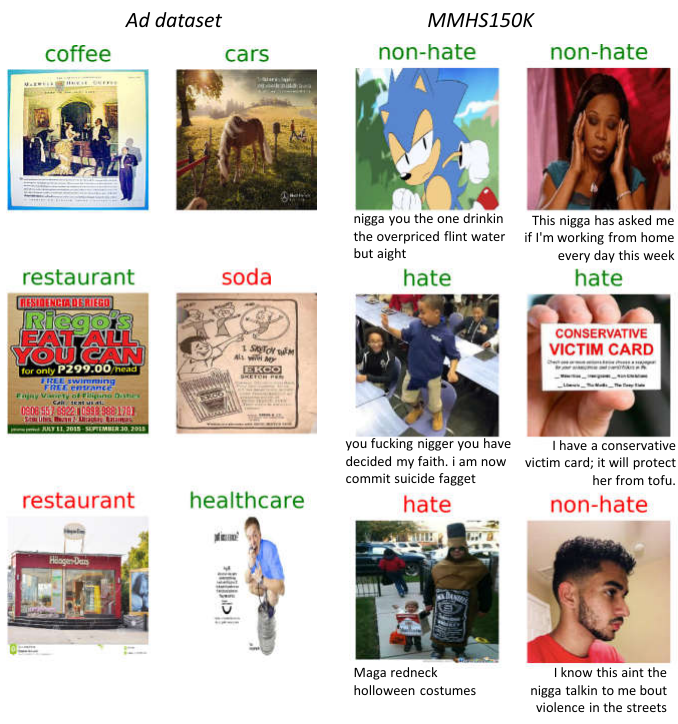}
\caption{ Qualitative Results on Classification task. Correct class labels are marked in green, and incorrect ones are marked with red }\label{fig:cl_qua}
\end{center}
\end{figure}

\subsection{Task2: Classification}
\paragraph{Topic Classification}
In the Topic classification task the objective is to classify an Ad image into 1 of 38 Topic classes. Topic classification was initially attempted by \cite{ad_hussain2017automatic} by training 152-layer ResNet using the visual features only. 
In Tab.\ref{tab:result_class_comp} row 1, we see another scheme\cite{deydon} which uses both text and visual features, but uses simple concatenation to aggregate them. Thus, we stress that simple use of text is not enough; we have to find ways to capture the multimodal interaction, as in our contextual encoding. In  row 3, we observe that using features from the network trained on the sentence relevance task, we can already improve upon the previous results. In row 4, we present the results of our final end to end trained topic classifier.

\paragraph{Tweet Classification}
In the recently proposed Hate Speech dataset MMHS150K\cite{gomez2019exploring}, the binary classification task of marking tweets, containing both visual and text content, as "hateful" and "benign" was proposed. 
We address the class imbalance problem in the original dataset by training on equal number of random samples from each class. Application of our contextual encoder, leads to results comparable with the multimodal models proposed by the authors
\begin{table}
\begin{center}
\small
\caption{Classification results }\label{tab:result_class_comp}
\begin{tabular}{ll|ll}
\multicolumn{2}{l|}{Ads Dataset} & \multicolumn{2}{l}{MMHS150K} \\ \hline
Dey et al\cite{deydon}                                           & 58               \\
Hussain et al \cite{ad_hussain2017automatic}        & 64.34       & SCM\cite{gomez2019exploring}           & 68.5      \\
pretrained on Task1                                                 & 66.35     & TKM\cite{gomez2019exploring}            & 68.2      \\
\bf{Our Model trained}                              & \bf{69.23}       & \bf{Our Model trained}             & \bf{67.44}      \\
 \end{tabular}
 \end{center}
 \end{table}
\vspace{-5pt}
\section{Conclusion}\label{sec:conclusion}
\vspace{-7pt}
We proposed a framework for interpreting images by leveraging both visual and text contents present in the images. Visual cues, both symbols and objects, together with scene text are extracted and embedded in a semantic space trained with triplet loss. Our embedding also incorporates text-visual inter-object dynamics encoding, which leads to capturing non-literal relationships between the detected objects. This idea of extracting and encoding, followed by embedding in semantic space, finds application in semantic retrieval and classification tasks. \\
In addition, we leverage the linguistic structure, training separate branches of the network for action-reason parts of the statement. We augment the visual-text semantic representation of the image with the lexical similarity between scene text and the query statement. Results confirm our initial hypothesis that scene text plays an important role in semantic understanding of images. These results encourage us to extend the application of our framework to more generic domains, for instance, the recently released datasets\cite{tmg_singh2019textvqa,tmg_biten2019scene} for VQA using scene texts.  

\vspace{-5pt}
{\small
\bibliographystyle{ieee}
\bibliography{egbib2}
}

\end{document}